\documentclass{article}

\usepackage{PRIMEarxiv}

\usepackage[utf8]{inputenc} 
\usepackage[T1]{fontenc}    
\usepackage{hyperref}       
\usepackage{url}            
\usepackage{booktabs}       
\usepackage{amsfonts}       
\usepackage{nicefrac}       
\usepackage{microtype}      
\usepackage{lipsum}
\usepackage{fancyhdr}       
\usepackage{graphicx}       
\graphicspath{{media/}}     

\title{A Conceptual Algorithm for Applying Ethical Principles of AI to Medical Practice} 

\author{
    Debesh Jha$^{1}$, Gorkem Durak$^{1}$, Vanshali Sharma$^{1}$, Elif Keles$^{1}$, Vedat Cicek$^{1}$,\\
    Zheyuan Zhang$^{1}$, Abhishek Srivastava$^{1}$, Ashish Rauniyar$^{2}$, Desta Haileselassie Hagos$^{3}$,\\
    Nikhil Kumar Tomar$^{1}$, Frank H. Miller$^{1}$, Ahmet Topcu$^{4}$, Anis Yazidi$^{5}$,\\
    Jan Erik Håkegård$^{2}$, Ulas Bagci$^{1,*}$\\
    \\
    $^{1}$Machine \& Hybrid Intelligence Lab, Department of Radiology, Northwestern University, USA\\
    $^{2}$Sustainable Communication Technologies, SINTEF Digital, Norway\\
    $^{3}$Department of Electrical Engineering and Computer Science, Howard University, USA\\
    $^{4}$Department of General Surgery, Tokat State Hospital, Tokat-Türkiye\\
    $^{5}$OsloMet Artificial Intelligence (AI) Lab, Oslo Metropolitan University, Norway\\
    debesh.jha@northwestern.edu
}

\begin{document}
\maketitle

\begin{abstract}

Artificial Intelligence (AI) is poised to transform healthcare delivery through revolutionary advances in clinical decision support and diagnostic capabilities. While human expertise remains foundational to medical practice, AI-powered tools are increasingly matching or exceeding specialist-level performance across multiple domains, paving the way for a new era of democratized healthcare access. These systems promise to reduce disparities in care delivery across demographic, racial, and socioeconomic boundaries by providing high-quality diagnostic support at scale.
As a result, advanced healthcare services can be affordable to all populations, irrespective of demographics, race, or socioeconomic background. The democratization of such AI tools can reduce the cost of care, optimize resource allocation, and improve the quality of care. In contrast to humans, AI can potentially uncover complex relationships in the data from a large set of inputs and lead to new evidence-based knowledge in medicine. However, integrating AI into healthcare raises several ethical and philosophical concerns, such as bias, transparency, autonomy, responsibility, and accountability. In this study, we examine recent advances in AI-enabled medical image analysis, current regulatory frameworks, and emerging best practices for clinical integration. We analyze both technical and ethical challenges inherent in deploying AI systems across healthcare institutions, with particular attention to data privacy, algorithmic fairness, and system transparency. Furthermore, we propose practical solutions to address key challenges, including data scarcity, racial bias in training datasets, limited model interpretability, and systematic algorithmic biases. Finally, we outline a conceptual algorithm for responsible AI implementations and identify promising future research and development directions. 
\end{abstract}


\section{Introduction}
The integration of artificial intelligence (AI) is fundamentally transforming healthcare delivery across clinical practice, patient care, and healthcare system administration. AI technologies now enable high accuracy in medical diagnosis and precise treatment planning, while optimizing healthcare workflows and reducing medical errors. {\
color{blue}For patients, this translates into increasingly personalized care through AI systems that can interpret complex medical histories to generate tailored diagnostic, prognostic, and therapeutic recommendations~\cite{he2019practical}.}

\begin{figure} [!t]
    \centering
    \includegraphics[width =1 \textwidth]{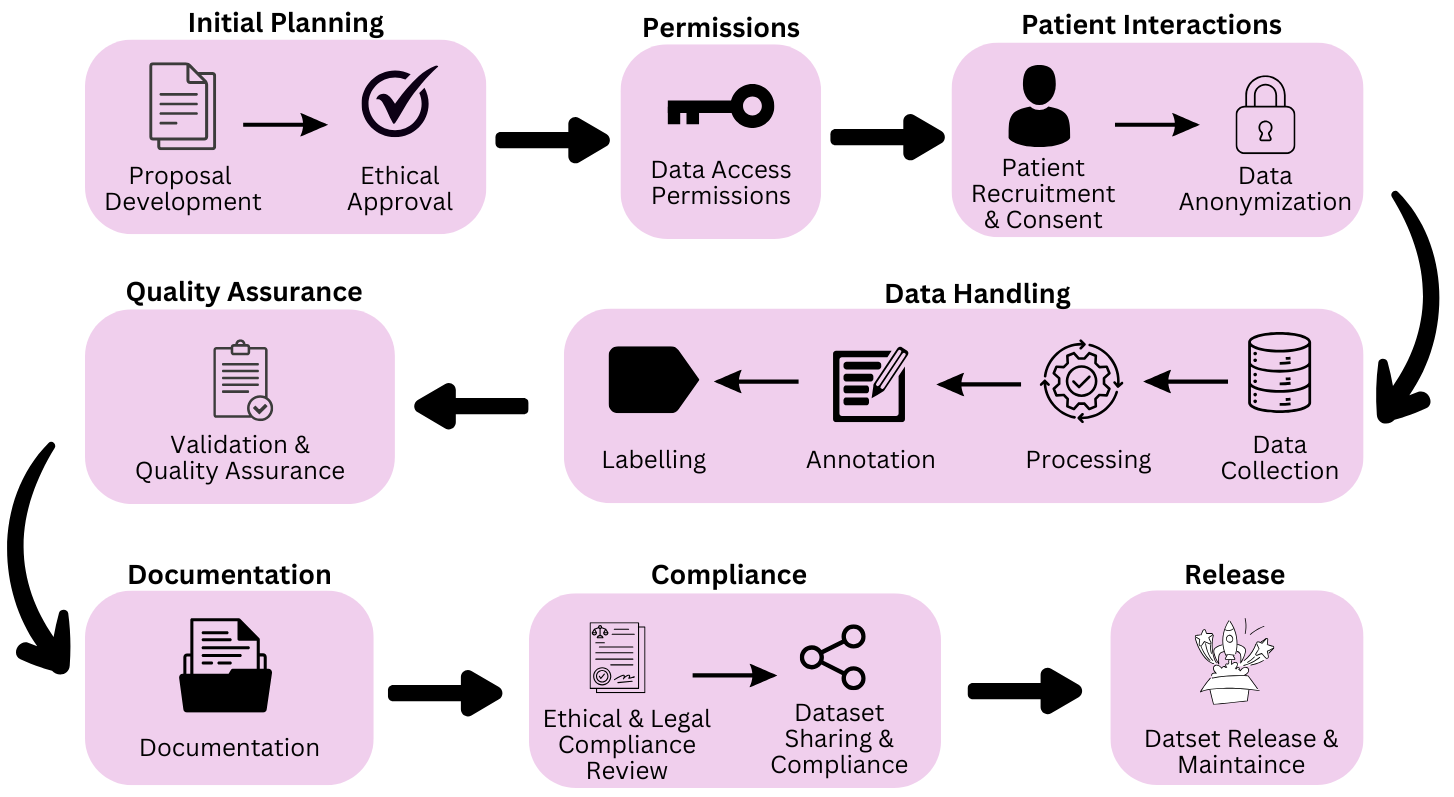}
    \caption{{The stepwise dataset development process.}}
    \label{fig:datadevelopment}
\end{figure}


The real-world impact of AI in healthcare is already evident across multiple domains. {In ophthalmology, the collaboration between Google and the Aravind Eye Care System achieved physician-level accuracy in detecting diabetic retinopathy~\cite{simonite2018google}, while uncovering previously unrecognized associations between fundus photographs and cardiovascular risk factors~\cite{poplin2018prediction}. This FDA-approved tool has proven particularly valuable in regions with ophthalmologist shortages. Similar breakthroughs include the FDA-approved GI Genius system for AI-assisted colonoscopy~\cite{medtronic2023} and significant advances in radiology, where AI enhances breast, lung, and prostate cancer screenings~\cite{mckinney2020international}. Commercial applications like the Koios DS for Breast and syngo.CT Lung CAD~\cite{fda2023breast} demonstrate AI's growing role in clinical diagnostics.}

{The healthcare industry's embrace of AI extends beyond diagnostics to practical clinical support tools. Google Health's Care Studio streamlines clinical workflows for conditions ranging from diabetes to hypertrophic cardiomyopathy~\cite{googlecarestudio2023}, while their \textit{DermAssist} app enables rapid skin condition assessment~\cite{googledermassist2023}.} These innovations, coupled with Medicare's approval of AI systems for medical image diagnosis, signal growing acceptance of AI in clinical care. Economic analyses suggest AI could generate annual savings of $150$ billion in the United States by 2026~\cite{collier2017artificial}.

\begin{table}[t!]
\caption{\label{tab:table1} Ethical and legal concerns and use of AI in different areas of medicine~\cite{safdar2020ethical,cacciamani2024artificial}. It is to be noted that  AI application, ethical concerns, and legal concerns can be for more than one `area of medicine'. }
\centering
\scriptsize
\begin{tabular}{|p{2.6 cm}|p{4.2cm}|p{2.8cm}|p{2.5cm}|}
\hline
\textbf{Areas of Medicine} &\textbf{AI uses} & \textbf{Ethical concern} & \textbf{Legal concern}\\
\hline
Diagnostic radiology & Clinical support system for examination of medical images  &Biased algorithms~\cite{Chau2024,Neri2020,Pesapane2018}  & Accountability~\cite{Contaldo2024,Neri2020,Pesapane2018}\\ \hline
Pathology & Improved accuracy & Biased decision~\cite{McKay2022,Niazi2019}  & Liability~\cite{McKay2022,Coulter2022}\\ \hline
Radiation oncology  & Improved treatment, reduced planning time, and radiotherapy & Lack of transparency~\cite{Huynh2020,Khanna2020}  & Data privacy and security~\cite{Khanna2021,Cohen2022}\\ \hline
Internal medicine & Electronic health record & Accessibility~\cite{Matsuzaki2018,McGreevey2020} & Intellectual property~\cite{Lupton2018,McGreevey2020} \\ \hline
Pediatrics & Predictive modeling & Bias~\cite{Keles2023,Coghlan2024} & Regulations~\cite{Keles2023,Desapriya2024}  \\ \hline
Surgery & Robot-assisted surgery & Safety~\cite{Naik2022,Snyder2024} & FDA regulations~\cite{Naik2022} \\ \hline
Emergency medicine & Medication management & Role of consent~\cite{Iserson2024} & Quality of care~\cite{Chenais2023}\\ \hline
Allergy and immunology & Risk identification &Impact of workforce~\cite{Khoury2022}  & HIPAA compliance~\cite{Goktas2024,Khoury2022,Khan2024,Breugel2023} \\ \hline
Prostate cancer & Drug optimization \& development  & Autonomy~\cite{Hesjedal2024,Agrawal2024,Goldenberg2019}  & Clinical trial regulations~\cite{Khanna2020,Khanna2021,Agrawal2024} \\ \hline
Urology& Identifying the most effective treatments  & Doctor-patient relationship~\cite{Cacciamani2024,Adhikari2024} & Malpractice~\cite{Cacciamani2024}  \\ \hline
\end{tabular}
\end{table}



{However, as AI revolutionizes healthcare across telemedicine, electronic health records, medical imaging, wearable devices, and robot-assisted surgery, it raises critical ethical considerations. This comprehensive analysis examines the ethical implications of AI-based healthcare technologies, providing structured guidelines for responsible deployment. We identify and address key challenges in data privacy, algorithmic fairness, clinical integration, and regulatory compliance, offering practical solutions for data scarcity, racial bias in training datasets, limited model interpretability, and systematic algorithmic biases. Through this analysis, we develop a conceptual algorithm for responsible AI implementation and outline promising directions for future research and development, ensuring that technological advancement aligns with ethical healthcare delivery and equitable patient access.}

\section{Ethical concerns of AI in medicine}
While AI technology has revolutionized healthcare through improved disease diagnosis, management, and treatment outcomes, it also presents significant ethical challenges that must be carefully addressed. These challenges span multiple dimensions, from fundamental ethical concerns to practical implementation issues and legal considerations.

\begin{figure}
    \centering
    \includegraphics[width=0.8\columnwidth]{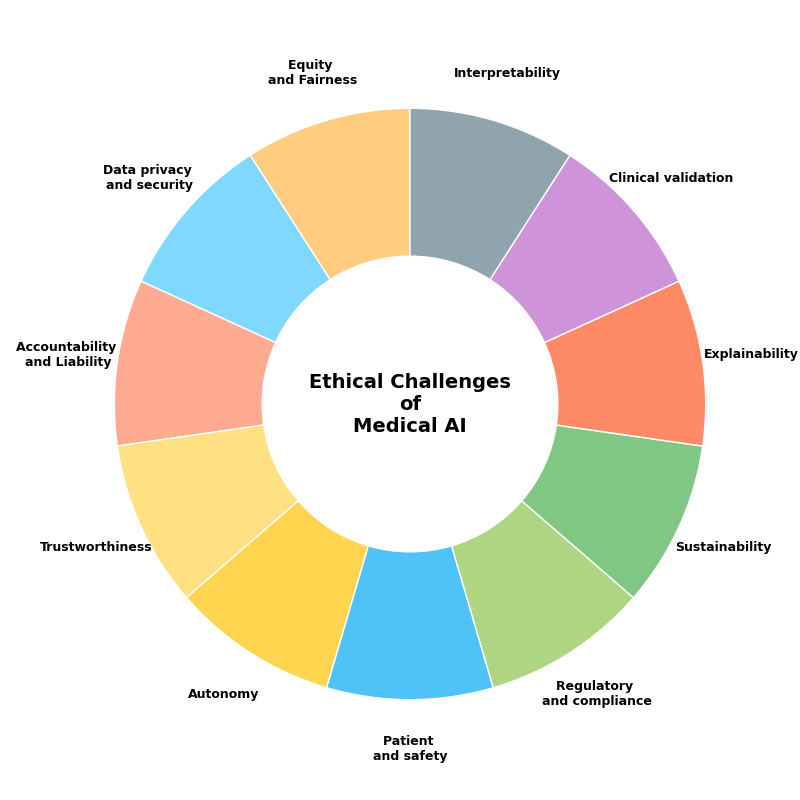}
    \caption{{Key ethical challenges in the Medical AI system.}}
    \label{fig:ethical_challenge}
\end{figure}

{The primary ethical concerns in medical AI include patient consent for data usage, privacy and transparency, algorithmic fairness, and data security~\cite{gerke2020ethical}. These intersect with legal considerations surrounding intellectual property rights, cybersecurity protocols, data protection laws, safety standards, and liability frameworks. Figure~\ref{fig:ethical_challenge} illustrates how these concerns extend to bias in decision-making, outcome explainability, system autonomy, interpretability, clinical validation, and broader issues of fairness and equity in healthcare delivery.}

Current ethical guidelines and policies have struggled to keep pace with rapid AI advancements in medicine. Despite efforts to engage the medical community~\cite{peek2015thirty,luxton2014recommendations}, healthcare professionals often lack a comprehensive understanding of the ethical complexities inherent in clinical AI implementation~\cite{rigby2019ethical}. Table~\ref{tab:table1} outlines these challenges, highlighting how ethical and legal concerns often overlap across different medical specialties. For instance, while algorithmic bias might be particularly visible in diagnostic radiology, similar challenges affect pathology, radiation oncology, and other medical fields.

{Addressing these challenges requires a multi-faceted approach focusing on:
\begin{itemize}
\item Clear communication of AI capabilities and limitations to both healthcare providers and patients.
\item Comprehensive training for clinicians on practical AI applications
\item Early involvement of healthcare professionals in AI tool development
\item Robust privacy and confidentiality protection measures
\item Well-defined roles for medical experts and AI systems
\item Structured protocols for informed consent and AI implementation
\end{itemize}
}

 The successful integration of AI in healthcare depends on fostering deeper understanding and trust among physicians and patients through continued research and open dialogue within the medical community. This approach ensures that technological advancement aligns with ethical healthcare delivery while maintaining patient safety and privacy as paramount concerns.

\section{Ethical datasets and algorithm development guidelines}
\subsection{Dataset guidelines}

\subsubsection{Dataset collection and construction}

The development of robust AI healthcare systems demands attention to dataset composition and collection methodology. Data collection should prioritize demographic diversity by incorporating cohorts from multiple clinical centers, ensuring comprehensive representation across population groups. This systematic approach is essential for developing generalizable deep learning models that perform consistently across diverse patient populations. {The data development process is shown in} Figure~\ref{fig:datadevelopment}. {It involves several steps, such as initial planning, where a proposal for data collection is written, and ethical approval is obtained.  Following this, data access permission is secured, and patient recruitment and consent are obtained. Data handling process is crucial as it contains important steps such as annotation and labeling. After this, quality assurance is performed to ensure data integrity. Then, comprehensive documentation for data is maintained. Then, we check for ethical and legal compliance. After this, we release the dataset. All these steps require strict adherence to legal and ethical standards for public domain usage, including systematic documentation of data sources, standardized collection protocols, verification of demographic representation, and compliance with privacy regulations. These structured protocols, combined with rigorous ethical review board approval and informed consent procedures, ensure that the algorithms exhibit minimal bias while maintaining high performance across diverse patient populations. Through this comprehensive data collection and validation approach, healthcare AI systems can better serve the full spectrum of patient needs while upholding the highest standards of ethical practice and clinical efficacy.}

\subsubsection{Ethical and privacy aspects of the data} 
{ The protection of patient privacy and ethical standards are fundamental requirements in healthcare dataset collection. This process requires a multi-layered approach to data security and ethical compliance. Written consent must be obtained from the participating medical centers, followed by a comprehensive review by data inspectors, local ethics committees, or institutional review boards. Before public release, datasets must undergo thorough anonymization to protect patient identities while preserving clinically relevant information. This rigorous framework ensures the development of valuable research resources while maintaining patient confidentiality and ethical integrity in the advancement of AI healthcare applications.}

\subsubsection{Addressing the data bias}
{ In most clinical studies, underrepresented patient subgroups may be overlooked (for example, populations like Black, Hispanic, Native American, or Native Hawaiian communities in the United States). Similarly, clinical studies often underrepresent diverse groups, including those based on disability, socioeconomic status, ethnicity, religion, gender orientation, age, linguistic capabilities, and multiple health conditions. Ignoring such underrepresented groups introduces bias to the algorithm. Thus, to prevent data bias, it is essential to include equal numbers of positive instances, negative instances, and rare cases so that the prediction made by the algorithm is correct. Therefore, to overcome the data bias problem, a transparent and inclusive data development process should be ensured before training the ML algorithms.} 

\subsubsection{High-quality dataset annotation}
{ Dataset annotation quality is critical for developing reliable ML models for healthcare applications. High-quality annotations, particularly at the pixel level for medical imaging, directly correlate with model performance and generalizability. Dataset developers have an ethical obligation to ensure precise annotations through standardized protocols and consistent tooling. To maintain annotation accuracy, rigorous quality control measures must be implemented, including thorough review processes and validation steps. This attention to detail in data labeling helps ensure that AI models learn appropriate features and maintain performance when deployed on new data distributions, ultimately supporting more reliable clinical applications.}

\subsubsection{Standard training, validation, and test split} 
{ Dataset developers must establish standardized training, validation, and testing splits to enable fair algorithm benchmarking and reproducible research. Official dataset partitioning serves multiple critical functions: ensuring consistent performance evaluation, reducing research redundancy, and enabling transparent comparison of competing methodologies. By providing these standardized splits, developers can facilitate streamlined progress in the field and eliminate discrepancies that arise from various data partitioning approaches. The implementation of mandatory official splits promotes research integrity and accelerates algorithmic advancement through clear and uniform evaluation standards.}

\subsection{Algorithm development guidelines}
\subsubsection{Effect of randomness} 

{ The integration of randomness in AI algorithms through weight initialization, data augmentation, dropout layers, and learning rate scheduling presents significant ethical challenges in healthcare applications. These stochastic elements can introduce unintended biases that affect patient care decisions and treatment recommendations, potentially creating disparities in model performance across different patient groups while reducing the interpretability of decision-making processes. The impact of algorithmic randomness extends beyond immediate clinical decisions. It affects model reproducibility and scientific validation, posing particular challenges for clinical trials and research verification where consistent, reproducible results are essential.}

To address these concerns, healthcare AI systems require rigorous monitoring of model performance across patient subgroups, clear documentation of randomness sources and their potential impacts, and transparent reporting of model uncertainty. Regular assessment and documentation of these random elements, combined with standardized protocols for reproducibility testing, are crucial for maintaining trust among healthcare providers and patients. These measures help ensure that AI systems maintain reliability and fairness in healthcare applications while benefiting from the advantages of stochastic methods. Through careful examination and continuous monitoring of randomness in model development and deployment, we can better safeguard patient interests while advancing AI capabilities in healthcare.


{ \subsubsection{Biased Algorithm} Algorithmic bias in healthcare AI systems manifests through multiple dimensions—race, gender, socioeconomic status, and even the unconscious biases of developers and stakeholders. These biases can have particularly severe consequences in healthcare, where they may perpetuate and amplify existing healthcare disparities. When AI systems display bias, they not only risk providing suboptimal care to marginalized communities but also erode trust in healthcare institutions, potentially deterring these populations from seeking necessary medical attention.}

Addressing algorithmic bias requires a comprehensive approach centered on model interpretability and data transparency~\cite{goodman2017european}. This involves several critical steps:
\begin{itemize}
\item Rigorous evaluation of training data representation and quality
\item Continuous monitoring and assessment of model decisions across different demographic groups
\item Implementation of interpretability techniques that make AI decision-making processes transparent to healthcare providers
\item Documentation of model limitations and potential biases
\end{itemize}
Before clinical deployment, healthcare providers must have access to detailed information about the model's training data, decision-making processes, and known limitations. This transparency enables clinicians to make informed decisions about AI tool implementation while considering their patient population's specific needs. Furthermore, bias detection and mitigation should be integrated throughout the development process, not treated as a one-time validation step.


{ \subsubsection{Training strategy and reporting hyperparameters} Deep Learning (DL) algorithms use hyperparameters, such as the learning rate, number of epochs, loss function, optimizer, batch size, learning rate schedule, regularization, normalization layers, model complexity, scaling factors, dropout rate, kernel size, and stride in convolutional layers. However, most studies did not report detailed hyperparameters. Thus, networks may exhibit significant variation in any medical image analysis task if the training strategy is varied. Furthermore, different researchers may use different training strategies, which may lead the method to achieve an unfair (increase/decrease) performance. Therefore, it is the ethical obligation of authors to document all relevant and minute steps essential for the reproducibility of results while developing DL models. In cases where space limitations preclude the inclusion of important details in the paper, authors can provide comprehensive information related to preprocessing, data partitioning, evaluation metrics, hyperparameters, and ethical considerations in publicly accessible repositories, such as GitHub.}

\subsubsection{Reproduciblity}
\begin{figure} [!t]
    \centering
    \includegraphics[width =1\textwidth]{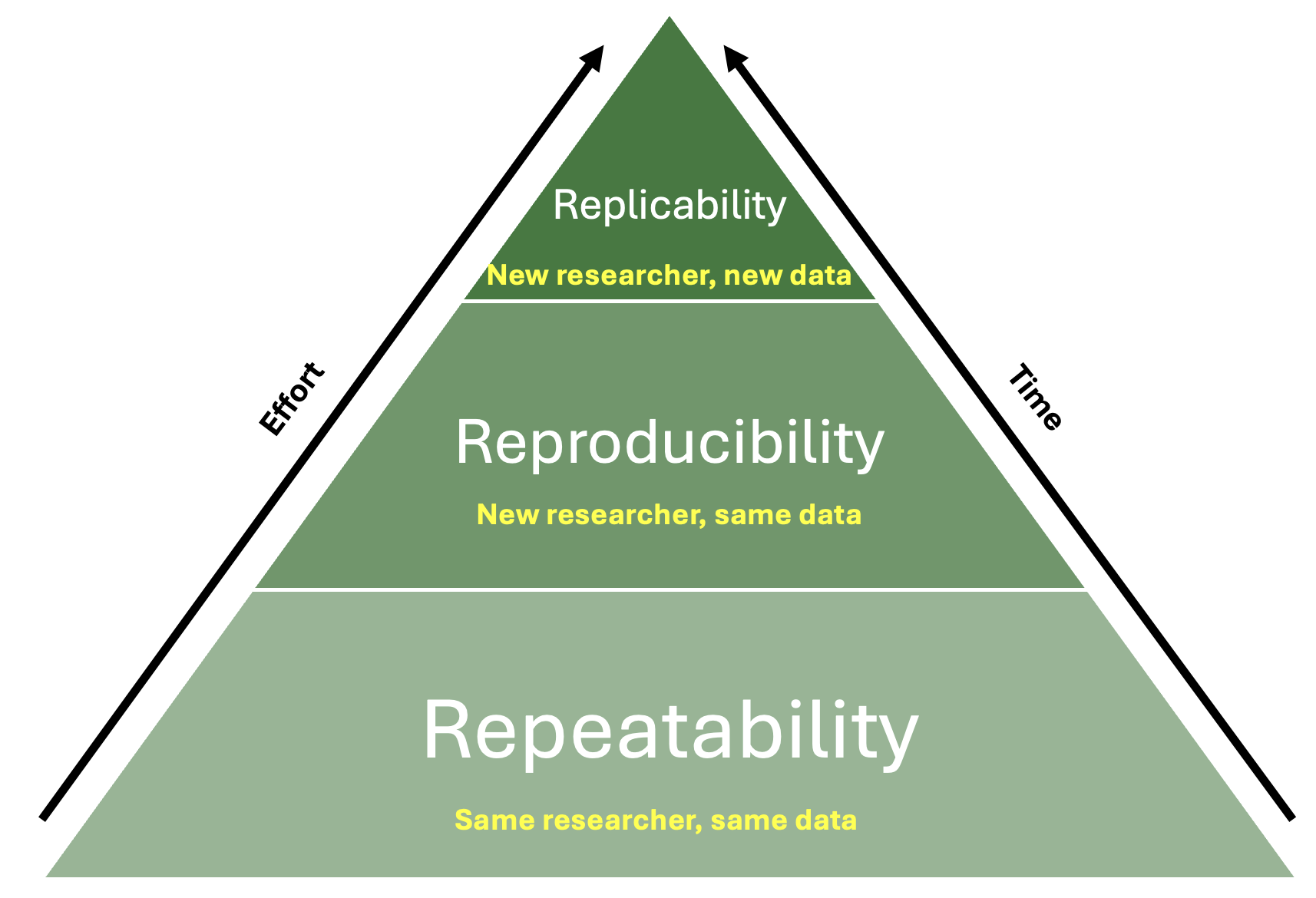}
    \caption{The problem of repeatability, reproducibility, and replicability.}
    \label{fig:my_label2}
\end{figure}

Algorithm developers are responsible for diligently scrutinizing the data distribution for potential biases, providing well-documented code, and offering a comprehensive account of the experimental setup. Figure~\ref{fig:my_label2} highlights the missing reproducibility in machine learning. Enhancing transparency and reproducibility can be achieved by sharing the source code and trained model weights, enabling fellow researchers to replicate and advance the work. This commitment to transparency and collaboration is an ethical imperative in algorithm development.

\subsection{Generalizability}
{ The lack of generalizability is a significant obstacle to the deployment of algorithms into clinical settings. This issue arises from the medical data collected from various centers and institutions that suffer from high variability. The differences in resolution, contrast, and signal-to-noise ratio is clear in medical data acquired using different imaging protocols. Thus, data from different imaging centers have different distributions. Such variations are insignificant when the trained AI model is tested on data from a distribution that the model already seen during training. However, the performance of the same AI system drops when it encounters data from an unseen distribution, therefore making its clinical application difficult. Thus, researchers should include generalization experiments in their research to address this issue. This can lead to healthy discussions and progress toward increased generalization of AI systems in medical image analysis, eventually facilitating its integration into healthcare systems.}

\subsubsection{Algorithm comparison}
{ An algorithm comparison using different evaluation metrics and diverse datasets is required to demonstrate the strengths and weaknesses of the existing and developed ML models against potential bias. This also ensures that the model is fair, robust, effective, and efficient in achieving the stated goals.} 

\subsection{Computational power} 
It is essential to report the computation power used in training and evaluating DL models. The information, such as the graphical processing units (GPU) model, software, and hardware used, can be useful during model evaluation. The information, such as the number of parameters in a model, CPU cores, memory, and the time it takes to train and evaluate the model, can be useful for understanding its complexity. Similarly, reporting the model's speed can provide a sense of the resources required to use the model in practice. Another important aspect to consider is the environmental impact of AI research, specifically, the greenhouse emissions and carbon footprint resulting from high energy consumption during training. This can help researchers make more sustainable choices in future research. {This information can be valuable for the reproducibility and benchmarking of different models and can help others understand the computation requirements of a given model.}

\subsection{Standard evaluation metrics for the task}
{ Agreeing on a standard set of metrics for medical image analysis tasks enables fair comparisons and facilitates the sharing of novel methodologies performance within the community. The research committee can also benefit from performing additional unnecessary experiments and saving computational resources. In some cases, the algorithm developer must conduct additional experiments and report the performance of additional metrics. Overall, the agreement of researchers over a standard set of metrics for certain types of tasks increases the pace of research and the ease of reviewing new methodologies.}

\subsection{Reporting limitation of research} 
{ Reporting limitations and listing open issues of research are essential when developing new algorithms. The limitations of the research can be the nature of the data and the poor generalization capability of the model, where the image sources come from different scanners or modalities. The developer can identify whether the algorithm predicts well for a single data type and fails for rare classes. This will motivate other researchers to solve the problem in open areas. Transparency in reporting limitations can also help build trust, demonstrating the researcher's commitment to creating ethical algorithms that can be integrated into clinics.}

\section{Towards solving key ethical challenges in Medical AI}
\subsection{Ethical consideration in data collection, use, sharing, and privacy} 
{ We need a considerable amount of data to develop the AI algorithm. We need a diverse dataset from multi-center institutions, including multiple scanners, cohort populations, and different resolutions, to develop robust and generalizable AI algorithms. Therefore, it is important to establish guidelines for dataset use and sharing. The data should be anonymized, de-identified, and approved by the patients before sharing~\cite{he2019practical}. The approval of data might be a signed informed consent before data collection and sharing. The \textit{data must be de-identified} by removing patients' identification information, such as names, addresses, and other personal details. Secure data-sharing platforms with appropriate safeguards can ensure data privacy and only authorized individuals can access patient data. Additionally, \textit{data governance frameworks} can be utilized to ensure that data is used ethically and responsibly. By following these steps, we can gain patients' trust and use their data to develop healthcare solutions.}

\subsection{Transparency}
{ Transparency is a crucial ethical challenge when deploying and developing medical AI tools. Most AI technologies are developed behind closed doors. Transparency is not only about the algorithm or data transparency but also about economic equality, authority, and governance. There are several ways in which transparency can be compromised. First, the DL models are complex and challenging to interpret.

In the context of medical imaging, for example, AI-driven radiology systems have shown promise in interpreting chest radiographs using deep learning models, enhancing diagnostic accuracy, and reducing radiologist workload~\cite{iqbal2024deep}. However, ethical concerns remain regarding data privacy, algorithmic transparency, and biases caused by imbalanced training datasets. Ensuring explainability and maintaining human oversight are essential to foster trust and accountability in such systems. Second, the AI system may make unfair or unjust decisions if the training data is biased. In addition, many commercial medical AI systems are proprietary, meaning that their inner workings are not available for inspection. This can make an understanding of how a system makes decisions, and whether it is doing so fairly can make it difficult. In addition, some companies or researchers may conduct experiments on the real-world deployment of AI models without informing users or may not share the results with the public, which may raise concerns about transparency. To address these challenges, developers can take several steps to increase transparency. First, the developer should consider \textit{interpretability to ensure fairness} and ensure that the data are unbiased while developing algorithms. Additionally, \textit{open-source systems} allow others to inspect the code and understand how the system works (See emergency of resource based theory~\cite{barney}). Moreover, \textit{sharing information} about the real-world deployment of the AI model and the user-friendly explainability of the AI model in decision-making can help gain trust. Furthermore, developers should consider a human-centric design approach to ensure transparency.}

\subsection{Explainability of AI}
\begin{figure}
    \centering
    \includegraphics[width=0.8\linewidth]{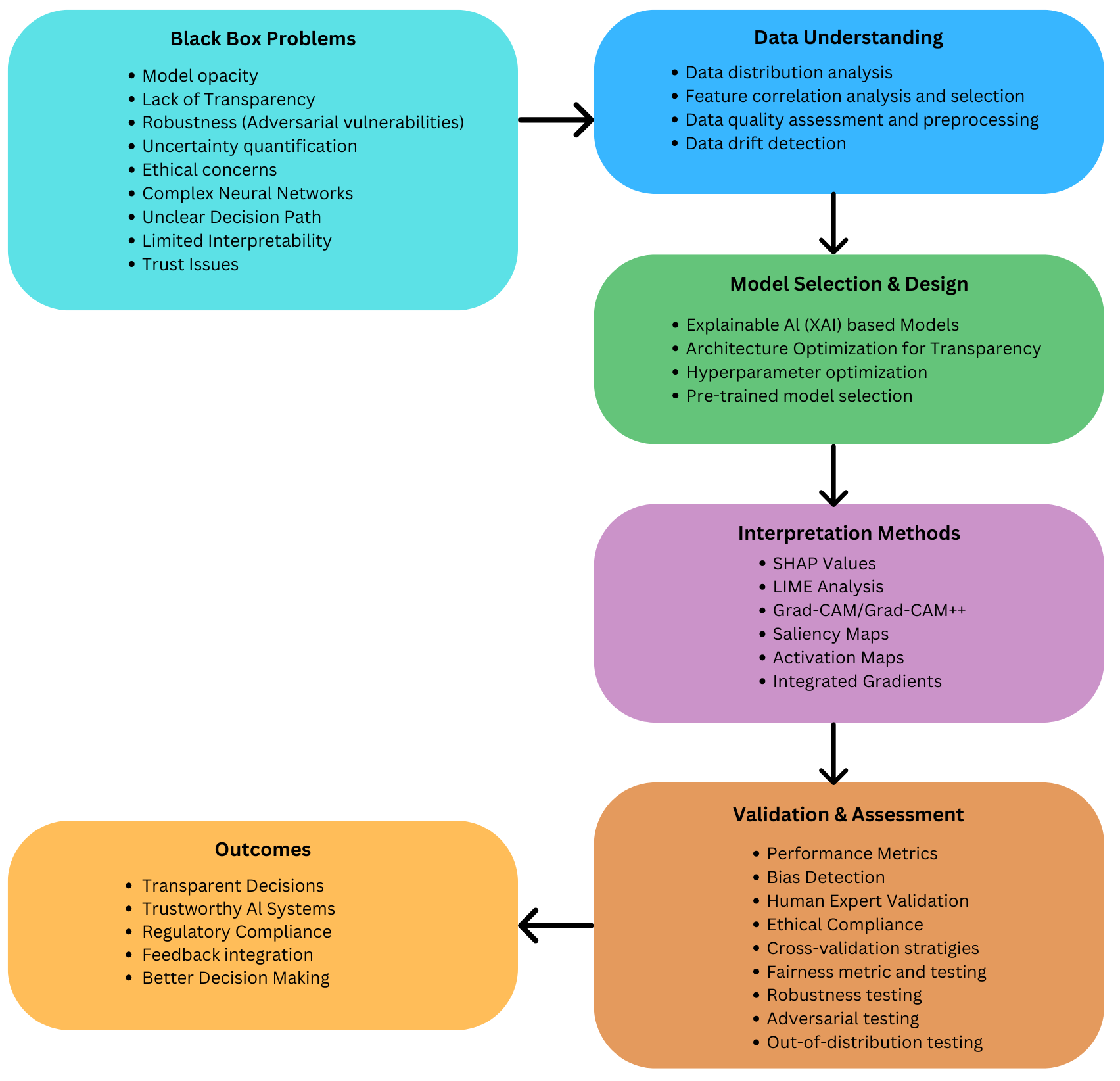}  
    \caption{{Black box problem and potential steps to achieve explainable AI.}}
    \label{fig:blackbox}
\end{figure}

{Explainability is a major ethical concern when developing medical AI systems. It can be subdivided into the `\textit{Black box problem}' and the '\textit{Clever Hans phenomenon}'. }

{
\textbf{The Black Box problem}: Figure~\ref{fig:blackbox} shows key issues associated with the black box problem and how we can overcome them. It refers to the difficulty of understanding the model prediction (for example, disease diagnosis). The lack of interpretability can make it challenging to evaluate the performance of such AI models and identify when they are making mistakes. Examples of black box problems include model opacity, lack of transparency, robustness issues, uncertainty quantification, ethical concerns, complex neural networks, unclear decision paths, limited interpretability, and trust issues.  This can make it challenging to hold healthcare organizations accountable for using the model. There are approaches to addressing this `Black box' problem. The model interpretability techniques can provide insights into how a model makes predictions. In addition, explainable AI (XAI) methods create models that produce human-understandable explanations for their decisions. Moreover, the decision surface of the model should be investigated, which is the boundary in the input space that separates the different output classes.  To address the black box problem, in the figure, we outline a structured approach to handle this with steps in data understanding, model selection and design, interpretation methods, validation and assessment, and desired outcomes. Approaches such as \textit{human in the loop'} systems as illustrated in the validation and assessment section, where experts actively participate in model evaluation and refinement, and \textit{transparency reporting' } as illustrated in the Outcomes,  which shows the model's limitations and decision pathways, can play a crucial role in a black box problem.}

\textbf{The Clever Hans phenomenon}: 
The risk of spurious correlations in medical AI systems require rigorous validation protocols. Deep learning models may develop diagnoses or treatment recommendations based on irrelevant data patterns, potentially leading to incorrect clinical decisions and patient harm. Proper evaluation through cross-validation tests, out-of-sample testing, and standardized metric assessment help identify and mitigate these spurious associations. These validation protocols are particularly crucial in clinical trials, where false positives can compromise research integrity and patient safety. Regular performance monitoring and systematic evaluation ensure AI systems maintain reliable clinical decision support based on medically relevant features rather than incidental correlations.

\subsection{Patient safety}
The regulation of AI-based healthcare technologies presents unique challenges distinct from traditional medical equipment oversight. Quality control mechanisms must evolve to address novel safety considerations, as highlighted by FDA recommendations for updated regulatory frameworks~\cite{he2019practical}. Implementing a human-in-the-loop approach ensures expert validation of AI predictions before clinical application, providing an essential safety layer in patient care.

Successful deployment requires continuous monitoring and performance evaluation in clinical settings. Regular system updates should incorporate new data and healthcare provider feedback to maintain optimal functionality and safety standards. All medical AI systems must strictly comply with established healthcare regulations, including HIPAA and GDPR requirements, to protect patient privacy and ensure data security. This comprehensive approach to quality control combines human oversight, continuous monitoring, and regulatory compliance to safeguard patient interests while advancing AI integration in healthcare.

\subsubsection{Accountability and Responsibility}
{ Accountability is also one of the main issues related to healthcare-based AI technology, and it is directly related to patient safety and trust. True transparency entails providing clear and accessible information from the \textit{initial design of the model, data source, deployment, and post-deployment processing}. Patients and healthcare professionals can better understand the model's capabilities and potential drawbacks and decide whether to use it. To mitigate accountability issues, a few actions can be taken, such as improving AI accuracy on representative and diverse datasets, agreements with patients, adopting policies, and regular verification of models by clinicians~\cite{choudhury2022impact}. If there is a false diagnosis, it should be brought to the \textit{software developer's attention } and considered an area of improvement. The \textit{participating physician, software developer, and third-party} involved in supplying should be made responsible. Clinicians should \textit{validate the outcomes of the AI models} to avoid any misleading diagnosis. \textit{Embedding safety and ethical protocol} when designing the medical AI system is important. Achieving accountability requires collaboration between \textit{academics, medical personnel, philosophers of ethics, and policy-makers}.}

\subsubsection{Overcome automation complacency}
{ Practitioners generally view the use of AI technology for diagnosis with skepticism, although automation complacency occasionally occurs. In this situation, physicians uncritically follow decisions predicted by automated systems. This may be attributed to the overwhelming cognitive workload that requires physician attention. This can lead to errors or inaccuracies in the AI model's predictions and can be a significant concern regarding patient safety.} Therefore, it is necessary to minimize automation complacency by monitoring automated tasks with proper attention~\cite{merritt2019automation}. Moreover, \textit{educating and training healthcare workers} about the AI model's capabilities and constraints and the significance of continual maintenance, monitoring, and critical review of the model's predictions is crucial. Instead of depending exclusively on the AI model's output, healthcare practitioners should be encouraged to critically assess the model's judgments and challenge them when is required.

\subsection{Ethical consideration when using AI-based algorithm}
The increasing sophistication of AI algorithms in healthcare, particularly reinforcement learning and meta-learning approaches, introduces complex ethical considerations. Healthcare organizations implementing AI-enabled medical systems must maintain clear accountability for algorithmic decisions that impact patient care. Regular evaluation of AI system performance across diverse populations is essential to identify and address potential biases or disparities in care delivery. {Resource-limited scenarios, such as the development of treatments for rare diseases and orphan drugs, further underscore the ethical challenges in medical AI. While AI can accelerate drug discovery in such contexts~\cite{irissarry2024using}, risks such as inequitable access and profit-driven biases must be mitigated through fairness, transparency, and regulatory oversight.} Before clinical deployment, developers must conduct thorough ethical reviews, ensuring that their systems uphold patient safety and equitable treatment standards while optimizing medical outcomes. This careful balance between algorithmic optimization and ethical compliance is crucial for responsible AI implementation in healthcare settings.

\subsection{Safeguarding against Malicious intent and use}
{ Malicious use of AI can involve altering the data used to train an AI system to achieve a specific outcome, biasing the model's predictions, creating fake data, or manipulating data to cause the AI system to make a wrong or harmful decision. Malicious use can also include using AI systems for purposes other than medicine or financial benefit by misusing patient or healthcare organizations' data. To be used in clinical settings, just beyond developing state-of-the-art AI algorithms, `productizing' is very difficult~\cite{dreyer2017machines}. To address ethical challenges, healthcare-based AI technologies should be committed to \textit{following key issues and correct guidelines} that might be raised by malicious AI use in clinical settings~\cite{toret2018}. Robust security mechanisms and protocols, including encryption, authentication, and access controls, must be implemented to secure sensitive health-related data to prevent AI's harmful use in healthcare settings. The system can also be detected for any malicious activities or vulnerabilities \textit{through regular system testing, monitoring, and updates}. A solid regulatory and supervisory structure is also essential to guarantee the safe and ethical use of AI applications in healthcare. The risk of malicious AI use in clinical settings can also be reduced by \textit{increasing awareness about the misuse of AI in the healthcare industry, training healthcare staff, and encouraging them to be watchful regarding suspicious activity and ensuring that these activities are promptly addressed}.}

\section{Ethical guidelines for  medical AI model deployment}

\begin{figure}[t!]
    \centering
    \includegraphics[width=\textwidth]{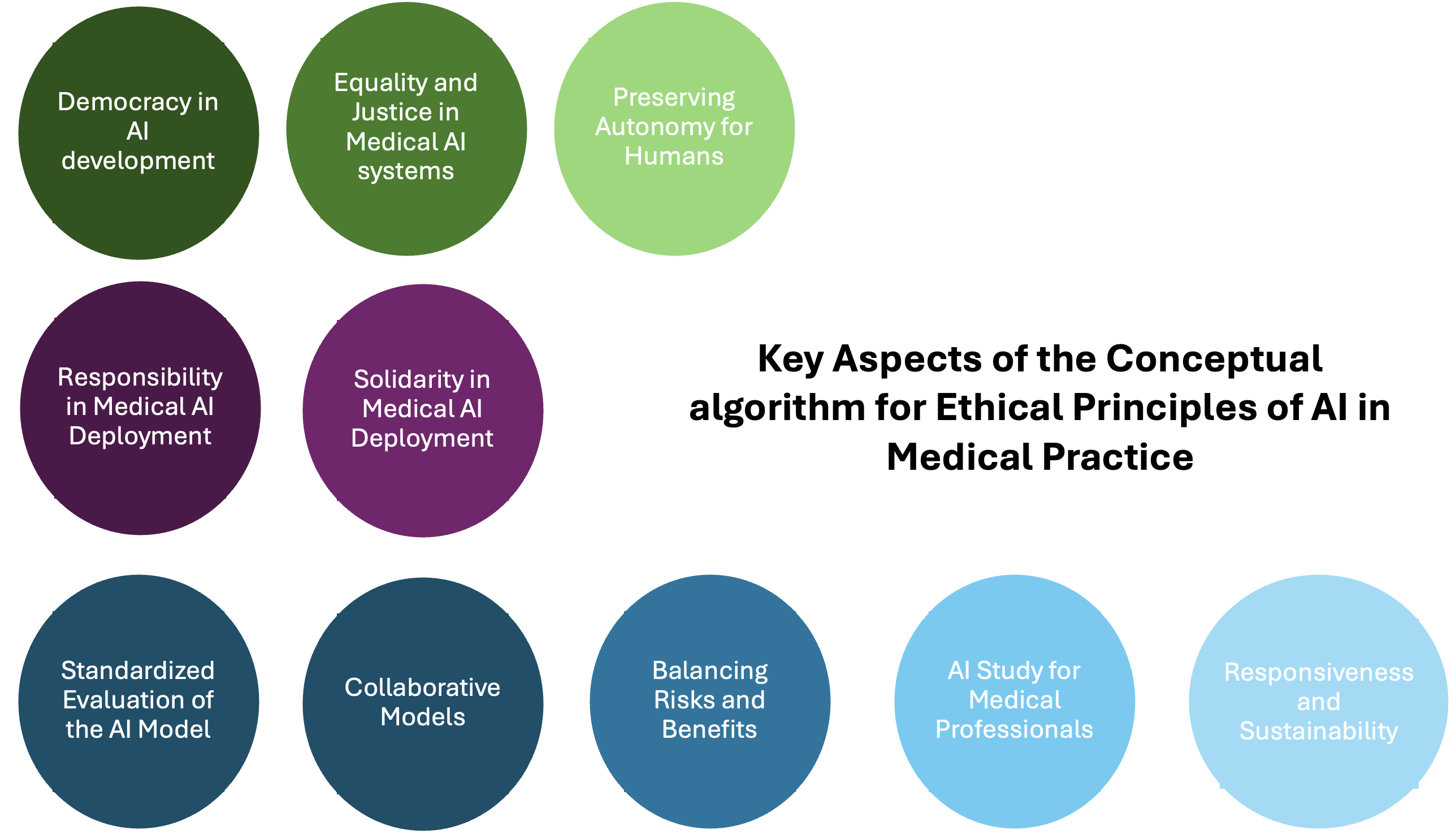}
    \caption{Key aspects of the conceptual algorithm for ethical principles of medical AI.}
    \label{fig:ethical_guidelines}
\end{figure}
The development of new AI healthcare technologies necessitates comprehensive ethical guidelines aligned with established medical principles, including the World Medical Association Declaration of Helsinki~\cite{world2001world,collier2017artificial}. Figure~\ref{fig:ethical_guidelines} {presents the key ethical guidelines important for building responsible AI development and deployment.} Below, we explain them in detail: 

\subsection{Preserving autonomy for humans} 
{ The integration of AI into healthcare must carefully balance technological advancement with patient autonomy. The fundamental principle that competent adults have the right to make informed decisions about their medical care remains paramount, even as AI systems increasingly inform clinical decision-making. While AI provides valuable insights, the ultimate authority for treatment decisions must remain with patients and their healthcare providers, who will never default to automated systems. Healthcare providers must retain the ability to override machine recommendations when clinically appropriate while ensuring robust data protection measures.}

For nations newly adopting healthcare AI, establishing comprehensive regulatory frameworks is essential. These frameworks must uphold World Health Organization guidance emphasizing human autonomy in healthcare decisions while implementing rigorous patient data protection protocols~\cite{guidance2021ethics}. Such measures ensure that technological advancement enhances rather than diminishes patient autonomy in medical decision-making.

\subsection{Democracy in AI development}
{ The development of ethical AI algorithms must be grounded on the democratic principles of fairness and equality. Healthcare AI systems require inclusive designs that ensure accessibility across gender, racial, and socioeconomic boundaries. Patients must be fully informed of the benefits and risks associated with AI-assisted diagnosis and treatment. System deployment should prioritize transparency and accountability so that stakeholders can understand and validate algorithmic decisions. These democratic principles extend beyond technical implementation and encompass equitable access to AI-enhanced healthcare services, regardless of demographic or economic factors.}

\subsection{Equality and justice in medical AI systems}
Medical data-driven AI systems use patient records, scans, and clinical history while making inferences. However, institutional biases toward disadvantaged groups and racial minorities are also reflected in such records. Therefore, the medical AI model trained on such data can lead to inductive biases toward disadvantaged, vulnerable patient groups~\cite{straw2020automation}. Moreover, unintended bias by medical professionals toward patients can also contribute to racial or gender-biased datasets. For example, clinicians have demonstrated a tendency to diagnose women with borderline personality disorder with the same symptoms as men diagnosed with post-traumatic stress disorder~\cite{becker1994sex}. In the realm of racial bias, AI-powered detection systems for skin cancer~\cite{guo2021bias,daneshjou2021lack} and retinal diagnostics~\cite{burlina2021addressing} have shown their proficiency in reporting high-performance for light-skinned individuals while failing to achieve decent performance for dark-skinned individuals. These inconsistencies stem from heavily imbalanced datasets used in training such methodologies, where the amount of data for light-skinned individuals exceeds that for dark-skinned individuals. Thus, the \textit{data acquisition protocol for training AI models must be regulated} to ensure equality and relevant publicly available datasets should be scrutinized. Researchers should also report the \textit{performance of their methods under different racial and gender classes to ensure the advancement of fair and trustworthy AI, devoid of racial, social, and gender biases}.

\subsection{Solidarity in medical AI deployment}
The rapid integration of artificial intelligence in healthcare—from robot-assisted surgery to automated diagnostics and treatment planning—represents a fundamental transformation of medical practice. While these advances promise improved care delivery, they also raise critical concerns about workforce displacement and economic equity. To ensure responsible implementation, comprehensive protection measures are essential across multiple domains. Healthcare institutions must establish robust reskilling programs for affected medical professionals, providing clear pathways for role evolution rather than elimination while preserving specialized medical expertise. Similarly, frameworks for fair compensation must recognize data contributors as key stakeholders and implement revenue-sharing models for those involved in ongoing data collection and AI system training.

Insurance industry practices require particular scrutiny, with mandatory transparency in AI-based policy decisions and explicit prohibitions against using genetic or historical data to deny coverage. Regular audits of AI systems used in policy determination should be standard practice. To oversee these various aspects, dedicated regulatory bodies must be established with clear guidelines for AI deployment in healthcare, mechanisms for protecting patient rights and privacy, and enforcement capabilities for ethical compliance. These coordinated measures would help ensure that technological advancement in healthcare benefits all stakeholders while protecting individual rights and professional dignity. The establishment of such comprehensive oversight mechanisms is crucial for maintaining public trust and ensuring { the equitable distribution of AI benefits across the healthcare ecosystem.}

\subsection{Responsiveness and sustainability in clinic}
{ Clinical AI systems must prioritize transparency, responsiveness, and sustainability in their design and deployment. Clear visibility of the decision-making processes can build trust between healthcare providers and patients. Regular monitoring and updates ensure systems stay aligned with the advancement of medical knowledge and AI capabilities, with swift responses to any identified inefficiencies. The environmental impact of AI deployment requires careful consideration, favoring lightweight, efficient models that minimize energy consumption and carbon footprint. Developers must balance high performance with environmental consciousness and economic accessibility to ensure AI healthcare benefits are provided to all communities. These priorities require an ongoing assessment of system adequacy, regular maintenance protocols, and commitment to sustainable development practices in healthcare AI implementation.}

\subsection{Responsibility in medical AI deployment}
As artificial intelligence systems become increasingly prevalent in clinical settings, establishing clear lines of responsibility emerges as a critical priority. While AI offers revolutionary capabilities, it also introduces new challenges and potential errors that require careful management. Legal and policy frameworks must clearly define the roles and responsibilities of healthcare providers, system developers, and vendors. The implementation of "human warranty" protocols ensures proper oversight of AI system development and clinical usage while established redressal mechanisms enable swift responses to adverse events. These frameworks must include clear procedures for investigating incidents, implementing corrective measures, and maintaining rigorous standards of patient care. Healthcare institutions can better harness AI's benefits while protecting patient interests through these structured approaches to responsibility and accountability.

\subsection{Standardized evaluation of the model}

The rapid advancement of AI in healthcare has created intense competition, with numerous claims of state-of-the-art performance lacking rigorous validation. This competitive environment risks compromising critical factors: privacy safeguards, deployment responsibility, algorithmic fairness, and transparent limitation reporting.

To maintain scientific integrity, we propose mandatory protocols:
\begin{itemize}
    \item Open-source code release,
    \item Validation on standardized test datasets,
    \item Reproducibility verification,
    \item Comprehensive fairness assessments,
    \item Transparent performance metrics,
    \item Documented limitation disclosure,
    \item Ethical design compliance.
\end{itemize}
{Such standardization accelerates innovation by establishing clear benchmarks while preserving research integrity. This framework allows researchers to focus on advancement rather than methodology validation to promote faster, more reliable progress in AI healthcare applications.

The implementation of these protocols will provide a robust foundation for comparing genuine advances while ensuring that ethical considerations remain paramount in development.}

\subsection{Collaborative models for medical AI system}
The European Union's strategy on trustworthy AI emphasizes the critical importance of collaboration between developers, stakeholders, policymakers, and domain experts in healthcare AI development~\cite{smuha2019eu}. This collaborative framework enables the integration of diverse perspectives, ensuring ethical compliance and fairness in AI healthcare solutions. By combining multidisciplinary expertise with standardized guidelines for collaboration, organizations can develop AI systems that better serve patient needs while maintaining ethical standards. This approach not only enhances system trustworthiness but also promotes more equitable and effective healthcare delivery through AI implementation.

\subsection{Balancing risks and benefits for medical AI system}
A thorough evaluation of AI integration in medicine requires careful assessment of both benefits and risks. While AI can enhance clinical efficiency through improved decision-making and workflow optimization, implementation challenges warrant careful consideration. The high initial investment requirements, potential workforce impacts, and limitations in replicating physician intuition and creativity must be weighed against operational benefits. Privacy concerns, patient consent requirements, and trust issues present additional challenges.

{The responsible deployment of AI in healthcare demands an emphasis on transparency and explainability. Patients should understand that AI systems primarily serve as screening tools and second opinions supporting physician judgment. Software developers must prioritize interpretable results and communicate system limitations to all stakeholders—doctors, algorithm developers, and patients. This transparency helps resolve technical uncertainties and builds trust in AI-augmented healthcare delivery.

AI systems in healthcare must maintain rigorous privacy protections to ensure successful integration while demonstrating clear clinical value. Regular updates and improvements are essential for maintaining system relevance and effectiveness over time. Success requires balancing technological capability with ethical considerations and practical implementation challenges.}

\subsection{AI study for medical professionals} 
{ The growing influence of AI in healthcare requires a better understanding of machine learning and deep learning fundamentals by medical professionals. This knowledge will empower physicians to evaluate AI-based studies and participate meaningfully and effectively in clinical trials. The combination of domain expertise and AI literacy allows medical professionals to provide more accurate data interpretation and valuable iterative feedback for system improvement. This interdisciplinary knowledge helps bridge the communication gap between AI developers and healthcare practitioners, ensuring the development of ethically sound and clinically beneficial models. When medical professionals understand both the technical and clinical aspects, they can better guide AI development to maintain high ethical standards while maximizing patient care benefits.}

\section{Discussion}
{The integration of AI into medicine requires careful consideration of ethical and social implications throughout the entire development pipeline, from conception to clinical deployment}. Each stage—design, development, and implementation—requires adherence to rigorous ethical guidelines and standardized protocols.

Our analysis reveals systematic biases in AI algorithms that can disadvantage marginalized communities, particularly in diagnostic and treatment recommendations. We propose specific corrective measures for implementation during the development and testing phases, including:
\begin{itemize}
    \item Diverse and representative training datasets
    \item Regular bias audits
    \item Validation across demographic groups
    \item Community engagement in development
\end{itemize}
Transparency is a critical factor for successful clinical integration. We advocate for standardized data protocols prior to clinical implementation, including:
\begin{itemize}
    \item Uniform data collection and storage formats
    \item Public access to validated datasets and algorithms
    \item Clear documentation of model limitations
    \item Standardized reporting of algorithm performance across population subgroups
\end{itemize}
These measures promote accountability while facilitating broader validation and improvement of AI systems in healthcare settings.

The clinical deployment of AI systems raises fundamental questions about human autonomy, automation bias, and professional accountability. Our investigation reveals a complex landscape of attitudes toward AI-assisted healthcare. While some clinicians and patients demonstrate high acceptance of AI recommendations—sometimes leading to {an uncritical reliance on automated systems—others express concerns about diminishing human agency in medical decision-making and the importance of individualized care.

The question of accountability for AI-assisted diagnostic errors remains particularly challenging. Current frameworks do not clearly define the responsibility of healthcare providers, AI system developers, and institutions. This uncertainty can undermine patient trust and willingness to engage with AI-enhanced care delivery. To address these challenges, we propose}:
\begin{itemize}
    \item Preserving patient autonomy through informed choice in treatment approaches
    \item Establishing clear accountability frameworks for AI-assisted decisions
    \item Strengthening clinician-patient communication regarding AI system capabilities and limitations
    \item Developing protocols to prevent automation bias while maintaining AI benefits
    \item Ensuring mechanisms for case-specific modifications to AI recommendations
\end{itemize}
These measures can help balance technological advancement with patient-centered care while maintaining appropriate human oversight in clinical decision-making.

As discussed in this article, the universal concerns related to AI in medicine need a standardized solution. For example, the International Medical Device Regulators Forum (IMDRF) provided guidelines for a common framework and defined software as a medical device (SaMD)  for regulators~\cite{imdrf2013software}. GDPR by the European Union provides guidelines for collecting, storing, and using the information at a personal level~\cite{goodman2017european}.  The American Medical Association's policy for healthcare-based AI technology only calls for responsive, designed-based, high-quality, and clinically approved AI technology.  Only this type of strategy for a medical system can assist as a prototype~\cite{rigby2019ethical}. Such policies are imperative to integrate ethical norms with AI technologies. {Additionally, there is a need for in-depth discussion between technologists, patients, and physicians. It can help gain trust among physicians, patients, and policymakers and give a sense of the acceptability of new tools.}

\section{Conclusion and Future Directions}

AI presents transformative opportunities in healthcare, including disease diagnosis, surveillance, drug development, and treatment protocols. However, its integration into medical practice demands careful consideration of potential impacts on human autonomy and established ethical frameworks. This article examines the philosophical and ethical implications of AI adoption in medicine, analyzing both its promise and potential pitfalls in current and future healthcare scenarios.

{The evidence} from medical centers and industry implementations demonstrates compelling efficacy for AI systems in clinical settings. However, this integration presents complex ethical challenges that require careful navigation. Critical concerns include protection of patient privacy, algorithmic transparency, data security, systematic bias, professional accountability, and safeguards against system misuse.

Beyond merely identifying these challenges, {we provide a comprehensive analysis} of potential solutions and implementation strategies. Our recommendations encompass frameworks for clinical AI deployment and ethical guidelines for dataset curation and algorithm development. These guidelines emphasize:
\begin{itemize}
    \item Robust privacy protection mechanisms
    \item Transparent algorithm validation processes
    \item Strategies to identify and mitigate bias
    \item Clear accountability structures
    \item Safeguards against misuse
    \item Protocols for ongoing monitoring and assessment
\end{itemize}
This balanced approach acknowledges AI's tremendous potential while establishing pragmatic frameworks to address ethical concerns, ultimately promoting responsible innovation in healthcare delivery.

{ The ethical guidelines provide insight into future directions for solving AI-related challenges in medicine. The process of employing AI, from the development stage to the deployment stage, should consider human rights and ethical standards as the core part to align with human-centric values. All stakeholders should be aware of ethical issues and should strictly follow the guidelines and principles recommended by the World Medical Association~\cite {world2001world}. Medical ethics~\cite{hansson2009philosophy} should be updated, and medical equipment based on AI should be monitored. The primary objective of AI in healthcare is to serve physicians and patients. Emphasis should be given to continuous research, open dialogue, and interdisciplinary collaboration to scrutinize social concerns, potential benefits, and risks associated with AI. Healthcare-based AI technology should be included in medical professionals' curricula to facilitate their participation and work with computer scientists and lawyers. All stakeholders should be included in the discovery, and the law should be enforced to make them responsible for coping with potential future problems. By putting all of these into action, we believe healthcare-based AI systems will eventually follow all ethical standards and be able to be used in various healthcare applications.}

\section*{Acknowledgments}
This project is supported by NIH funding: R01-CA246704 and R01-CA240639. This research work is also supported in parts by the COPS (Comprehensive Privacy and Security for Resilient CPS/IoT) project funded by the Research Council of Norway under project number: 300102.

\bibliographystyle{unsrt}  
\bibliography{references}

\begin{thebibliography}{10}

\bibitem{he2019practical}
Jianxing He, Sally~L Baxter, Jie Xu, Jiming Xu, Xingtao Zhou, and Kang Zhang.
\newblock The practical implementation of artificial intelligence technologies in medicine.
\newblock {\em Nature medicine}, 25(1):30--36, 2019.

\bibitem{simonite2018google}
T~Simonite.
\newblock Google’s ai eye doctor gets ready to go to work in india. wired magazine. june 8, 2017, 2018.

\bibitem{poplin2018prediction}
Ryan Poplin, Avinash~V Varadarajan, Katy Blumer, Yun Liu, Michael~V McConnell, Greg~S Corrado, Lily Peng, and Dale~R Webster.
\newblock Prediction of cardiovascular risk factors from retinal fundus photographs via deep learning.
\newblock {\em Nature Biomedical Engineering}, 2(3):158, 2018.

\bibitem{medtronic2023}
Medtronic.
\newblock Gi genius™ intelligent endoscopy module, 2023.
\newblock Accessed: 2023-09-16.

\bibitem{mckinney2020international}
Scott~Mayer McKinney, Marcin Sieniek, Varun Godbole, Jonathan Godwin, Natasha Antropova, Hutan Ashrafian, Trevor Back, Mary Chesus, Greg~S Corrado, Ara Darzi, et~al.
\newblock International evaluation of an ai system for breast cancer screening.
\newblock {\em Nature}, 577(7788):89--94, 2020.

\bibitem{fda2023breast}
U.S. Food and Drug Administration.
\newblock computer-assisted diagnostic software for lesions suspicious for cancer, 2023.
\newblock Accessed: 2023-09-16.

\bibitem{googlecarestudio2023}
Care Studio.
\newblock Clinical software to unify healthcare data, 2023.
\newblock Accessed: 2023-09-16.

\bibitem{googledermassist2023}
Google Health.
\newblock Dermassist, 2023.
\newblock Accessed: 2023-09-16.

\bibitem{collier2017artificial}
Matthew Collier, Richard Fu, Lucy Yin, and P~Christiansen.
\newblock Artificial intelligence: healthcare’s new nervous system.
\newblock {\em AI: Healthcare's new nervous system}, 2017.

\bibitem{safdar2020ethical}
Nabile~M Safdar, John~D Banja, and Carolyn~C Meltzer.
\newblock Ethical considerations in artificial intelligence.
\newblock {\em European journal of radiology}, 122:108768, 2020.

\bibitem{cacciamani2024artificial}
Giovanni~E Cacciamani, Andrew Chen, Inderbir~S Gill, and Andrew~J Hung.
\newblock Artificial intelligence and urology: ethical considerations for urologists and patients.
\newblock {\em Nature Reviews Urology}, 21(1):50--59, 2024.

\bibitem{Chau2024}
M.~Chau.
\newblock Ethical, legal, and regulatory landscape of artificial intelligence in australian healthcare and ethical integration in radiography: A narrative review.
\newblock {\em Journal of Medical Imaging}, 2024.

\bibitem{Neri2020}
E.~Neri, F.~Coppola, V.~Miele, C.~Bibbolino, and R.~Grassi.
\newblock Artificial intelligence: Who is responsible for the diagnosis?
\newblock {\em Radiologia Medica}, 125:517--521, 2020.

\bibitem{Pesapane2018}
F.~Pesapane, C.~Volonté, M.~Codari, and F.~Sardanelli.
\newblock Artificial intelligence as a medical device in radiology: ethical and regulatory issues in europe and the united states.
\newblock {\em Insights into Imaging}, 9:745--753, 2018.

\bibitem{Contaldo2024}
M.T. Contaldo, G.~Pasceri, G.~Vignati, and L.~Bracchi.
\newblock Ai in radiology: Navigating medical responsibility.
\newblock {\em Diagnostics}, 14(14):1506, 2024.

\bibitem{McKay2022}
F.~McKay, B.J. Williams, and G.~Prestwich.
\newblock The ethical challenges of artificial intelligence-driven digital pathology.
\newblock {\em The Journal of Pathology: Clinical Research}, 2022.

\bibitem{Niazi2019}
M.K.K. Niazi, A.V. Parwani, and M.N. Gurcan.
\newblock Digital pathology and artificial intelligence.
\newblock {\em The Lancet Oncology}, 2019.

\bibitem{Coulter2022}
C.~Coulter, F.~McKay, and N.~Hallowell.
\newblock Understanding the ethical and legal considerations of digital pathology.
\newblock {\em The Journal of Pathology: Clinical Research}, 2022.

\bibitem{Huynh2020}
E.~Huynh, A.~Hosny, C.~Guthier, and D.S. Bitterman.
\newblock Artificial intelligence in radiation oncology.
\newblock {\em Nature Reviews Clinical Oncology}, 2020.

\bibitem{Khanna2020}
S.~Khanna, S.~Srivastava, and I.~Khanna.
\newblock Ethical challenges arising from the integration of artificial intelligence (ai) in oncological management.
\newblock {\em Journal of Responsible AI}, 2020.

\bibitem{Khanna2021}
S.~Khanna, I.~Khanna, and S.~Srivastava.
\newblock Ai governance framework for oncology: Ethical, legal, and practical considerations.
\newblock {\em Quarterly Journal of Clinical and Translational Health}, 2021.

\bibitem{Cohen2022}
E.B. Cohen and I.K. Gordon.
\newblock First, do no harm. ethical and legal issues of artificial intelligence and machine learning in veterinary radiology and radiation oncology.
\newblock {\em Veterinary Radiology \& Ultrasound}, 2022.

\bibitem{Matsuzaki2018}
T.~Matsuzaki.
\newblock Ethical issues of artificial intelligence in medicine.
\newblock {\em California Western Law Review}, 2018.

\bibitem{McGreevey2020}
J.D. McGreevey, C.W. Hanson, and R.~Koppel.
\newblock Clinical, legal, and ethical aspects of artificial intelligence-assisted conversational agents in health care.
\newblock {\em JAMA}, 2020.

\bibitem{Lupton2018}
M.~Lupton.
\newblock Some ethical and legal consequences of the application of artificial intelligence in the field of medicine.
\newblock {\em Semantic Scholar}, 2018.

\bibitem{Keles2023}
E.~Keles and U.~Bagci.
\newblock The past, current, and future of neonatal intensive care units with artificial intelligence: a systematic review.
\newblock {\em npj Digital Medicine}, 6:1--12, 2023.

\bibitem{Coghlan2024}
S.~Coghlan, C.~Gyngell, and D.F. Vears.
\newblock Ethics of artificial intelligence in prenatal and pediatric genomic medicine.
\newblock {\em European Journal of Human Genetics}, 2024.

\bibitem{Desapriya2024}
E.~Desapriya, P.~Tiu, and C.~Ma.
\newblock Overlooked benefits, risks, and assumptions of ai integration in pediatric care.
\newblock {\em JAMA Pediatrics}, 2024.

\bibitem{Naik2022}
N.~Naik, B.M.Z. Hameed, D.K. Shetty, and D.~Swain.
\newblock Legal and ethical consideration in artificial intelligence in healthcare: who takes responsibility?
\newblock {\em Frontiers in Surgery}, 2022.

\bibitem{Snyder2024}
K.B. Snyder, R.A. Stewart, and C.J. Hunter.
\newblock Ethical considerations for the application of artificial intelligence in pediatric surgery.
\newblock {\em Journal of Pediatric Surgery}, 2024.

\bibitem{Iserson2024}
K.V. Iserson.
\newblock Informed consent for artificial intelligence in emergency medicine: A practical guide.
\newblock {\em American Journal of Emergency Medicine}, 2024.

\bibitem{Chenais2023}
G.~Chenais, E.~Lagarde, and C.~Gil-Jardin\u00e9.
\newblock Artificial intelligence in emergency medicine: viewpoint of current applications and foreseeable opportunities and challenges.
\newblock {\em JMIR Medical Informatics}, 2023.

\bibitem{Khoury2022}
P.~Khoury, R.~Srinivasan, S.~Kakumanu, and S.~Ochoa.
\newblock A framework for augmented intelligence in allergy and immunology practice and research\u2014a work group report of the aaaai health informatics, technology, and augmented intelligence workgroup.
\newblock {\em Journal of Allergy and Clinical Immunology: In Practice}, 2022.

\bibitem{Goktas2024}
P.~Goktas and E.~Damadoglu.
\newblock Future of allergy and immunology: Is ai the key in the digital era?
\newblock {\em Annals of Allergy, Asthma \& Immunology}, 2024.

\bibitem{Khan2024}
M.~Khan, S.~Banerjee, S.~Muskawad, and R.~Maity.
\newblock The impact of artificial intelligence on allergy diagnosis and treatment.
\newblock {\em Current Allergy and Asthma Reports}, 2024.

\bibitem{Breugel2023}
M.~van Breugel, R.S.N. Fehrmann, M.~Bügel, and F.I. Rezwan.
\newblock Current state and prospects of artificial intelligence in allergy.
\newblock {\em Allergy}, 2023.

\bibitem{Hesjedal2024}
M.B. Hesjedal, E.H. Lysø, and M.~Solbjør.
\newblock How medical doctors, scientists and patients relate ethical challenges with artificial intelligence decision-making support tools in prostate cancer diagnostics to good outcomes.
\newblock {\em Sociology of Health \& Illness}, 2024.

\bibitem{Agrawal2024}
S.~Agrawal and S.~Vagha.
\newblock A comprehensive review of artificial intelligence in prostate cancer care: State-of-the-art diagnostic tools and future outlook.
\newblock {\em Cureus}, 2024.

\bibitem{Goldenberg2019}
S.L. Goldenberg, G.~Nir, and S.E. Salcudean.
\newblock A new era: artificial intelligence and machine learning in prostate cancer.
\newblock {\em Nature Reviews Urology}, 2019.

\bibitem{Cacciamani2024}
G.E. Cacciamani, A.~Chen, I.S. Gill, and A.J. Hung.
\newblock Artificial intelligence and urology: ethical considerations for urologists and patients.
\newblock {\em Nature Reviews Urology}, 2024.

\bibitem{Adhikari2024}
K.~Adhikari, N.~Naik, B.M.Z. Hameed, and S.K. Raghunath.
\newblock Exploring the ethical, legal, and social implications of chatgpt in urology.
\newblock {\em Current Opinion in Urology}, 2024.

\bibitem{gerke2020ethical}
Sara Gerke, Timo Minssen, and Glenn Cohen.
\newblock Ethical and legal challenges of artificial intelligence-driven healthcare.
\newblock In {\em Artificial intelligence in healthcare}, pages 295--336, 2020.

\bibitem{peek2015thirty}
Niels Peek, Carlo Combi, Roque Marin, and Riccardo Bellazzi.
\newblock Thirty years of artificial intelligence in medicine (aime) conferences: A review of research themes.
\newblock {\em Artificial intelligence in medicine}, 65(1):61--73, 2015.

\bibitem{luxton2014recommendations}
David~D Luxton.
\newblock Recommendations for the ethical use and design of artificial intelligent care providers.
\newblock {\em Artificial intelligence in medicine}, 62(1):1--10, 2014.

\bibitem{rigby2019ethical}
Michael~J Rigby.
\newblock Ethical dimensions of using artificial intelligence in health care.
\newblock {\em AMA Journal of Ethics}, 21(2):121--124, 2019.

\bibitem{goodman2017european}
Bryce Goodman and Seth Flaxman.
\newblock European union regulations on algorithmic decision-making and a “right to explanation”.
\newblock {\em AI magazine}, 38(3):50--57, 2017.

\bibitem{iqbal2024deep}
Hammad Iqbal, Arshad Khan, Narayan Nepal, Faheem Khan, and Yeon-Kug Moon.
\newblock Deep learning approaches for chest radiograph interpretation: A systematic review.
\newblock {\em Electronics}, 13(23):4688, 2024.

\bibitem{barney}
Jay~B Barney.
\newblock Why resource-based theory's model of profit appropriation must incorporate a stakeholder perspective.
\newblock {\em Strategic Management Journal}, 39(13):3305--3325, 2018.

\bibitem{choudhury2022impact}
Avishek Choudhury and Onur Asan.
\newblock Impact of accountability, training, and human factors on the use of artificial intelligence in healthcare: Exploring the perceptions of healthcare practitioners in the us.
\newblock {\em Human Factors in Healthcare}, 2:100021, 2022.

\bibitem{merritt2019automation}
Stephanie~M Merritt, Alicia Ako-Brew, William~J Bryant, Amy Staley, Michael McKenna, Austin Leone, and Lei Shirase.
\newblock Automation-induced complacency potential: Development and validation of a new scale.
\newblock {\em Frontiers in psychology}, 10:225, 2019.

\bibitem{irissarry2024using}
Carla Irissarry and Thierry Burger-Helmchen.
\newblock Using artificial intelligence to advance the research and development of orphan drugs.
\newblock {\em Businesses}, 4(3):453--472, 2024.

\bibitem{dreyer2017machines}
Keith~J Dreyer and J~Raymond Geis.
\newblock When machines think: radiology’s next frontier.
\newblock {\em Radiology}, 285(3):713--718, 2017.

\bibitem{toret2018}
Tore Tennøe.
\newblock {\em Artificial Intelligence – Opportunities, Challenges and a Plan for Norway}.
\newblock 2018.

\bibitem{world2001world}
World~Medical Association et~al.
\newblock World medical association declaration of helsinki. ethical principles for medical research involving human subjects.
\newblock {\em Bulletin of the World Health Organization}, 79(4):373, 2001.

\bibitem{guidance2021ethics}
WHO Guidance.
\newblock Ethics and governance of artificial intelligence for health.
\newblock {\em World Health Organization}, 2021.

\bibitem{straw2020automation}
Isabel Straw.
\newblock The automation of bias in medical artificial intelligence (ai): Decoding the past to create a better future.
\newblock {\em Artificial intelligence in medicine}, 110:101965, 2020.

\bibitem{becker1994sex}
Dana Becker and Sharon Lamb.
\newblock Sex bias in the diagnosis of borderline personality disorder and posttraumatic stress disorder.
\newblock {\em Professional Psychology: Research and Practice}, 25(1):55, 1994.

\bibitem{guo2021bias}
Lisa~N Guo, Michelle~S Lee, Bina Kassamali, Carol Mita, and Vinod~E Nambudiri.
\newblock Bias in, bias out: Underreporting and underrepresentation of diverse skin types in machine learning research for skin cancer detection—a scoping review.
\newblock {\em Journal of the American Academy of Dermatology}, 2021.

\bibitem{daneshjou2021lack}
Roxana Daneshjou, Mary~P Smith, Mary~D Sun, Veronica Rotemberg, and James Zou.
\newblock Lack of transparency and potential bias in artificial intelligence data sets and algorithms: a scoping review.
\newblock {\em JAMA dermatology}, 157(11):1362--1369, 2021.

\bibitem{burlina2021addressing}
Philippe Burlina, Neil Joshi, William Paul, Katia~D Pacheco, and Neil~M Bressler.
\newblock Addressing artificial intelligence bias in retinal diagnostics.
\newblock {\em Translational Vision Science \& Technology}, 10(2):13--13, 2021.

\bibitem{smuha2019eu}
Nathalie~A Smuha.
\newblock The eu approach to ethics guidelines for trustworthy artificial intelligence.
\newblock {\em Computer Law Review International}, 20(4):97--106, 2019.

\bibitem{imdrf2013software}
IMDRF SaMD~Working Group et~al.
\newblock Software as a medical device (samd): key definitions, 2013.

\bibitem{hansson2009philosophy}
Sven~Ove Hansson.
\newblock Philosophy of medical technology.
\newblock In {\em Philosophy of Technology and Engineering Sciences}, pages 1275--1300. 2009.

\end{thebibliography}

\end{document}